# Improving Image co-segmentation via Deep Metric Learning

**Zhengwen Li**
School of Computer Science,
Beijing Polytechnic University
lzw@bit.edu.cn

**Xiabi Liu**
School of Computer Science,
Beijing Polytechnic University
liuxiabi@bit.edu.cn

## Abstract

Deep Metric Learning (DML) is helpful in computer vision tasks. In this paper, we firstly introduce DML into image co-segmentation. We propose a novel Triplet loss for Image Segmentation, called IS-Triplet loss for short, and combine it with traditional image segmentation loss. Different from the general DML task which learns the metric between pictures, we treat each pixel as a sample, and use their embedded features in high-dimensional space to form triples, then we tend to force the distance between pixels of different categories greater than of the same category by optimizing IS-Triplet loss so that the pixels from different categories are easier to be distinguished in the high-dimensional feature space. We further present an efficient triple sampling strategy to make a feasible computation of IS-Triplet loss. Finally, the IS-Triplet loss is combined with 3 traditional image segmentation losses to perform image segmentation. We apply the proposed approach to image co-segmentation and test it on the SBCoseg dataset and the Internet dataset. The experimental result shows that our approach can effectively improve the discrimination of pixels' categories in high-dimensional space and thus help traditional loss achieve better performance of image segmentation with fewer training epochs.

## 1. Introduction

The loss function is one of the most important factors in the applications of deep learning, including image segmentation. Currently, the cross-entropy (CE) loss [1], Dice loss[2], and Focal loss[3] are the 3 most widely used loss functions in this field.

The existing image segmentation loss function mainly focuses on the distinguishability of low-dimensional semantic features of pixels, ignoring the discrimination of high-dimensional features of pixels from different categories. Theoretically, samples are more likely to be correctly distinguished in high-dimensional space than in low-dimensional space, but the similarity measurement problem in high-dimensional feature space needs to be solved.

Deep Metric Learning (DML) is concerned with learning a distance function tuned to a particular task and is useful when used in conjunction with nearest-neighbor methods and other techniques that rely on distances or similarities, *e.g.* face recognition [4-7], image retrieval [8-11]. In recent years, it has also attracted more and more attention in image segmentation research [12-15]. These works use the idea of Contrastive loss, and their efficiency and practicality need to be improved.

In this paper, we propose a novel Triplet loss function for Image Segmentation. We present a special strategy to sample foreground pixels and background pixels to construct triples. Then we force the distance between heterogeneous pixels greater than that of similar pixels in the high-dimensional feature space so that the pixels are easier to be distinguished in the high-dimensional feature space. We call this loss as IS-Triplet and combine it with classic image segmentation loss functions to perform image segmentation. We apply our proposed loss to image co-segmentation. The experiments are conducted on SBCoseg and Internet dataset. Our main contributions are summarized as follows:

(1) We propose a novel triplet loss function for image segmentation, called IS-Triplet loss for short. We treat each pixel in the image as a sample and train their high-dimensional embedding feature vector with our IS-Triplet loss to improve the distinguishability between foreground and background pixels. To our knowledge, this is the first introduction of triplet loss into image segmentation.

(2) A new triple sampling strategy is presented under our framework, which is important for the success of DML. We not only control the upper limit of the sampling number of the triples under the premise of ensuring the effect but also keep the weights of the previous foreground points and the background points in the triples, to prevent the adverse effects caused by the imbalanced number of the two types of pixels.

(3) We combine the proposed IS-Triplet loss with traditional image segmentation loss functions to perform image co-segmentation. The experiments on SBCoseg [16] and Internet dataset[17] prove that our loss can lead to better performance than traditional segmentation loss and this improvement can be observed stably. Although Dice loss, CE loss, and Focal loss have different performances, they can be improved to almost the same level after applying our method.

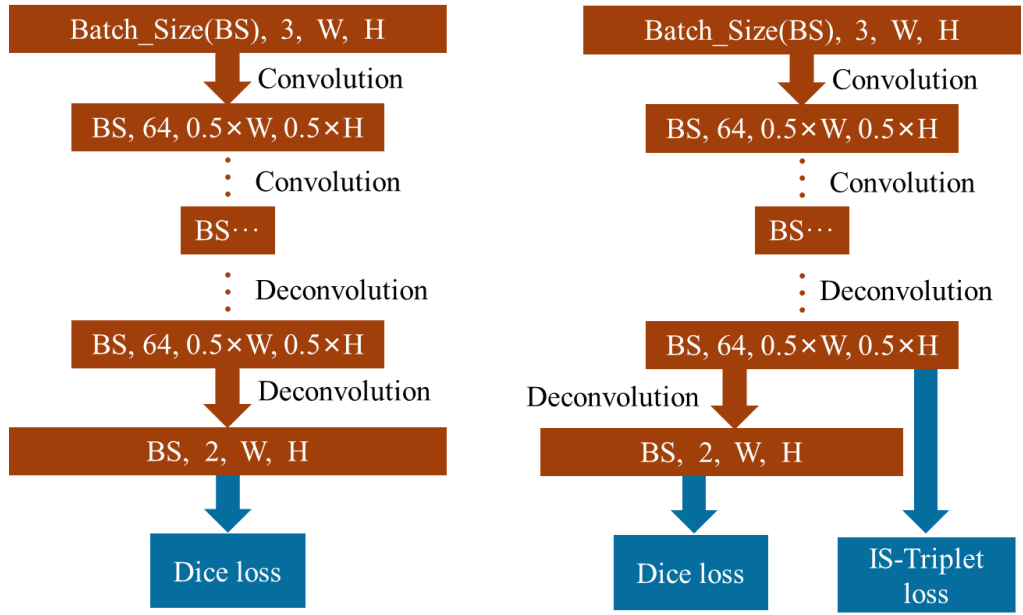

Fig. 1　A fully convolutional neural network is modified to apply our metric learning method. Where BS is the batch size, W and H are the width and height of the input image: (a) Schematic diagram of the original network structure; (b) Schematic diagram of the modified network structure.

# 2. Related Work

## 2.1. Loss Functions in Image Segmentation

The softmax outputs of the deep network for segmentation reflect the probabilities of classifying each pixel into the foreground class. The CE loss measures the quality of image segmentation from the view of single pixels. It aims at the correct classification of each pixel in the image [18-20] and ignores the integrity of the foreground or background objects. So, if most of the pixels have been classified correctly, the segmentation quality is not easy to be improved further. More importantly, the two classes are usually highly unbalanced in image segmentation. For example, the interesting foreground objects often occupy only a small part of the image. The CE loss is unsatisfactory to deal with such class unbalance problem.

To address the foreground-background class imbalance encountered during the training of dense detectors, Lin *et al.* [3] proposed Focal loss, which reshapes the standard CE loss such that it down-weights the loss assigned to well-classified examples. However, it still focuses on the classification of single pixels and ignores the integrity of foreground or background objects.

The Dice loss [21, 22], also known as intersection over union (IoU) or Jacarrd index, computes the overlap ratio between the ground-truth object region and the predicted object region to assess the segmentation quality. Because the integrity of foreground and background objects is considered, the Dice loss usually leads to better segmentation results than the CE loss.

## 2.2. Deep Metric Learning

Deep metric learning maps an image into a feature vector in a manifold space via deep neural networks. In this manifold space, the Euclidean distance or the cosine distance can be directly used as the distance metric between two points. Loss function also plays a key role in successful DML frameworks and a large variety of loss functions have been proposed in the literature. Contrastive loss [7, 8, 23, 24] captures the relationship between pairwise data points, *e.g.* similarity or dissimilarity.

Triplet loss [5, 25, 26] was first introduced by Florian *et al.* [5] in face recognition applications. Hoffer *et al.* [4] applied the method of triple loss specifically to metric learning to learn the feature embedding of data. A triplet is composed of an anchor point, a similar (positive) data point, and a dissimilar (negative) data point. The purpose of triplet loss is to learn a distance metric by which the anchor point is closer to a similar point than the dissimilar one by a margin. In general, the Triplet loss outperforms the contrastive loss, because the relationship between positive and negative pairs is considered.

After the triplet loss, more methods form multiple groups with more samples, such as quadruplet loss [27], N-pair loss [28], and lifted structured loss [11]. However, their core idea is still based on triplet loss, the final performance has not been significantly improved compared with triplet loss[29].

One of the main challenges in DML is the large sample space. For example, the number of triples for a triple network is $O(N^3)$ where N is the number of training set samples, it is usually impossible to exhaust all possibilities when training. To improve efficiency, various difficult sample mining methods [6, 9, 30-33] have been proposed.

In recent years, instance segmentation researchers have begun to pay attention to DML. Zhu *et al.* [12] learned a distance metric model from the atlases to automatic segmentation of the hippocampus from MR brain images. Liu *et al.* [15] proposed a DML enhanced neural network to promote the lesion segmentation result from the existing method. Fathi *et al.* [13] adopted the idea of contrastive loss [7, 8, 23, 24] to calculate the similarity between pairs of pixels so that the similarity of pixels from the same instance approaches 1 and that from different instances approaches 0. Chen *et al.* [14] used a fully convolutional network trained by a modified triplet loss as the embedding model to tackle the problem of video object segmentation.

Most DML tasks take each image as a sample and calculate the similarity between images. In this work, our goal is to distinguish the foreground and background points of an image. We propose a metric learning loss function to learn more discriminative convolution features for image segmentation. To the best of our knowledge, this is the first time that the idea of triplet loss has been applied to the field of image segmentation.

## 3. The Proposed Approach

We use DML to force the pixels from different categories more distinguishable in the high-dimensional feature space, thereby improving the classification effect in the low-dimensional semantic space. First, we regard each pixel in an image as a sample, so a certain number of triples can be formed by selecting foreground points and background points. By optimizing the triplet loss, the distance between the feature embedding of similar pixels goes smaller than that of heterogeneous pixels. Second, we present a targeted triple sampling method to improve computational efficiency. Finally, we combine the proposed method with the traditional image segmentation loss function to perform image co-segmentation.

### 3.1. Triplet loss for Image Segmentation

Although the metric learning method based on triplet loss[4, 25] was proposed very early, its core idea is simple and practical with good performance[29]. A triplet is composed of an anchor point, a similar (positive) data point, and a dissimilar (negative) data point. The purpose of triplet loss is to learn a distance metric by which the anchor point is closer to the similar point than the dissimilar one by a margin:

$$l_{triplet}(X_a, X_p, X_n; f) = [d_{ap}^2 + m - d_{an}^2]_+ \quad (1)$$

Where $X_a, X_p, X_a$ denotes the anchor point, positive point, and negative point, respectively. $f$ is the embedding function, $d_{ij}^2 = \|f(x_i) - f(x_j)\|_2$ is the Euclidean distance, $m$ is the violate margin that requires the distance of negative pairs to be larger than the distance of positive pairs, $[\cdot]_+$ is the hinge function.

Metric learning methods in image analysis usually treat an image as a sample. Differently, our method regards each pixel as a sample and the high-dimensional features corresponding to each pixel are used for calculation. In this work, we only study the problem of distinguishing the foreground and the background of an image, so only two categories are considered. As shown in Figure 1, when an image is inputted into the convolutional neural network, the output of each layer of the convolutional network contains position information of pixels. The size of an image's feature vector is first compressed and then enlarged until the size is the same as the input size after a certain layer of the network. It means that each pixel is represented by a D-dimensional feature vector. We use pixels to form triples and use the corresponding D-dimensional features to form triples. Then we use the IS-Triplet loss to force the feature distances of similar pixels to decrease, and the feature distances of heterogeneous pixels to increase.

## 3.2. A New Triple sampling strategy

The sampling strategy is an important part of DML. For a data set with a data volume of n, the order of magnitude of possible triples is $O(N^3)$, and most of the triples cannot provide effective information. In our applications to image segmentation, not only the number of images but also the number of pixels in each image are huge. It is neither possible nor necessary for us to optimize all possible triples. We propose an efficient triple sampling strategy to solve this problem.

Our method is to limit the total number of triples and balance the weights of the two types of pixels. In particular, as shown in Figure 2, we randomly select K foreground points from each image, use their multi-channel feature vectors to form a set F1, and repeat this operation to obtain another feature vector set F2. Note that some common pixels are allowed in F1 and F2, but their order in the two sets is generally different. Then we use the same method to form 2 feature vector sets B1 and B2 from the background points. If the number of foreground points or background points in an input picture is smaller than K, K takes the smaller value. Through the above operations, on one hand, we control the total number of triples to 2K, which effectively limits the amount of calculation; on the other hand, we can avoid imbalance problem in building triplets, even if the number of foreground pixels is different greatly from that of background pixels.

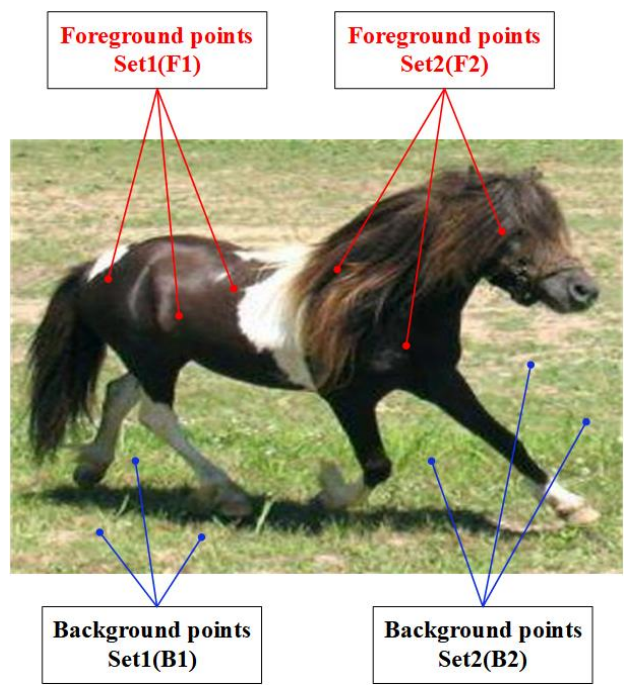

Fig. 2  Two sets of K pixels are randomly selected from foreground points or background points of an image

Now, we can use the embedded feature vector set F1, F2, B1, B2 to build 2K triples, which results in 2 part of loss:

$$loss1 = [d^2(F1, F2) - d^2(F1, B1) + m]_+ \quad (2)$$
$$loss2 = [d^2(B1, B2) - d^2(B1, F2) + m]_+ \quad (3)$$

where $d(A, B)$ denotes the Euclidean distance between the corresponding row vector of vectors A and B, $m$ is the violate margin that requires the distance of negative pairs to be larger than the distance of positive pairs, $[\cdot]_+$ is the hinge function.

The above Eq. 2 ensures that the embedding feature distance between any two foreground points is smaller than that between any one foreground point and any background point. Eq. 3 is used to ensure that the distance between the embedding features of any two background points is smaller than that between any one foreground point and any background point. Since the number of points in F1, F2, B1, and B2 are K, it is convenient to optimize 2K triples at once, thus ensuring that the distance between the similar pixels is shorter than the distance of the dissimilar pixels. Finally, we add loss1 and loss2 to form our IS-Triplet loss:

$$loss_{IS-triplet} = \frac{1}{2}([d(F1, F2) - d(F1, B1) + m]_+ + [d(B1, B2) - d(B1, F2) + m]_+) \quad (4)$$

## 3.3. Application to Image Co-Segmentation

We apply the proposed IS-Triplet loss to an image co-segmentation framework [34]. As shown in Figure 1, in the original framework [34], the output of the sixth deconvolution layer is a tensor with 2 channels and half the size of the input image. The Dice loss uses this 2-channel embedding to classify pixels in the semantic space. To apply our IS-Triplet loss, we made the following changes to the network structure. First, the output of the sixth deconvolution layer is modified to 64 channels and the size of the feature map is changed to be consistent with the size of the input image. This feature embedding is used in two places. One side is used for metric learning, and the other side is passed through a new convolutional layer with a kernel size of 1 to make it into 2 channels, with the same size kept, and then use a traditional loss to optimize as before. Finally, the two loss functions are combined by using

$$loss = loss_{seg} + \lambda loss_{IS-triplet} \quad (5)$$

Where $\lambda$ is a trade-off coefficient.

Our modified framework is shown in Figure 3. The corresponding deep network is obtained by embedding the correlation block[35] into a Siamese U-net network [34]. The overall architecture of the resultant network is composed of four parts. The first part is the Siamese encoders that are a pair of two feature encoder networks, each of which extracts the features from an image. The two encoders share the weights. We adopt the ResNet-50 network [36] to construct our Siamese encoder. The second part is the correlation block, through which the correlation maps are calculated from the two feature maps. The third part is the Siamese decoders, which are constructed by symmetrically reversing each of the two encoders and concatenating each scale of feature maps into the corresponding layers in the reversed pathway.

The core content of this paper and the improvement to the original framework are mainly reflected in the fourth part. We have modified the original network, using the latest feature embedding layer of the neural network with the same size as the input image, and using our IS-Triplet loss for metric learning. For more detailed information about this original co-segmentation framework and correlation block, please refer to[34].

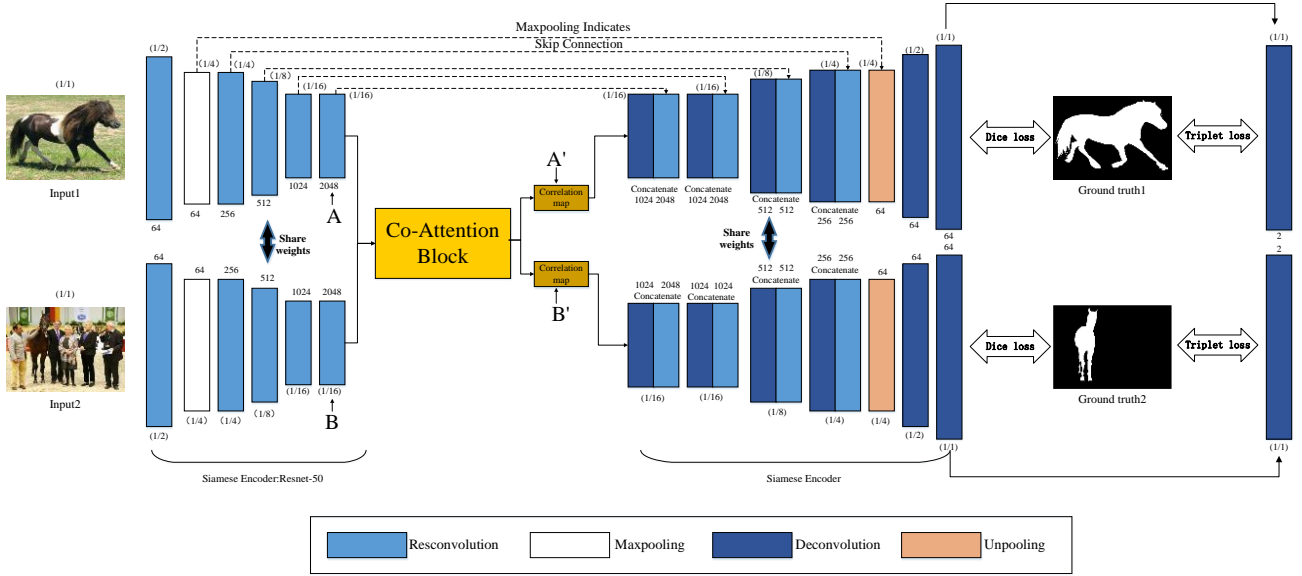

Fig. 3 The modified image co-segmentation network architecture.

## 4. Experiments

We designed 2 groups of experiments. In the first group, we combine IS-triplet loss with Dice loss and compared it with the use of only Dice loss. In the second group, the IS-triplet loss is used to combination with Dice loss, CE loss, and Focal loss to further study the effects of our approach. In the following sections, we describe our experiments in more detail.

## 4.1. Experimental Setup

### 4.1.1. Datasets

In the first group of experiments, we use the SBCoseg dataset[16], which includes 889 groups of images and each group consists of 18 images with a common object, leading to 16002 images in total. The whole dataset is divided into five subsets: with ECFB, with TR, with MH, with SD, and Normal (normal data). The five subsets contain 193, 251, 82, 83, and 280 image groups, respectively. Each original image is in JPG format with a pixel size of 360 ×360, and each ground-truth image is in PNG format.

In both of the two groups of experiments, we use the collection of Pascal VOC 2012[1][37] and MSRC[2][38] datasets to train our image co-segmentation network, we call it the collection dataset for short and then use the Internet [17] dataset as the test set. These three data sets are widely used in the community of image co-segmentation. MSRC is composed of 591 images of 21 object groups. The ground-truth is roughly labeled, which does not align exactly with the object boundaries. VOC 2012 includes 11,540 images with ground-truth detection boxes and 2913 images with segmentation masks. Only 2913 images with segmentation masks can be considered in our problem. Note that not all of the examples in these two datasets can be used. In MSRC, some images include only stuff without obvious foregrounds, such as only the sky or grassland. In VOC 2012, the interested objects in some images have great changes in appearance and are cluttered in many other objects, so that the meaningful correlation between them is ambiguous. We exclude them from consideration. The remained 1743 images in VOC 2012 and 507 images in MSRC are used to construct our training set. From the training images, we sampled 41,329 pairs of images containing common objects to train our proposed image co-segmentation network.

---

[1] Pattern analysis statistical modeling and computational learning, visual object classes, 2012.
[2] Microsoft Research Cambridge.

The Internet [17] dataset consists of 3 classes (airplane, car, and house) of thousands of downloaded Internet images. Following the compared methods, we evaluate our approach on its widely used subset, in which each class has 100 images.

### 4.1.2. Evaluation Metrics

We use two commonly used metrics for evaluating the effects of image co-segmentation: Precision and Jaccard index. Precision is the percentage of correctly classified pixels in both background and foreground, which can be defined as

$$\text{Precision} = \frac{|Segmentation \cap Ground\ truth|}{|Segmentation|}$$

Jaccard index (denoted by Jaccard in the following descriptions) is the overlapping rate of foreground between the segmentation result and the ground-truth mask, which can be defined as

$$\text{Jaccard} = \frac{Segmentation \cap Ground\ truth}{Segmentation \cup Ground\ truth}$$

### 4.1.3. Implementation Details

We conduct the experiments on a computer with RTX 2080Ti GPU and implement the image co-segmentation network with PyTorch[39]. In the experiments, the learning rate is initialized to 0.001 without IS-Triplet loss or 0.0001 with IS-Triplet loss, because we find these parameters perform best in the corresponding experiments. The learning rate is decreased to 0.85 times after each epoch. The weight decay and the momentum parameters are set to be 1e-4 and 0.9, respectively. The optimization procedure ends after 30 epochs.

For the setting of hyperparameters of our approach, the value of K is set to 5000, the violation boundary m is set to 3.0, and $\lambda$ is initialized to 1.0 and decreased to 0.85 times after each epoch.

In the image co-segmentation farmework, the input data are pairs of similar images, we matched the training data for the experiments in advance. Because of limited computing resources, when training on the Collection datasets, all images are resized to the resolution of 448×448 in advance. The batch size for training is set to be 3. The co-segmentation results are resized back to the original image resolution for performance evaluation.

## 4.2. The First Group of Experiments

To test the effectiveness of our proposed method, we combine IS-triplet loss with Dice loss and compared it with the only use of Dice loss. We conduct the above experiments on two different datasets. The first dataset is the relatively simple SBCoseg dataset [16], both the training set and the validation set come from this dataset, and there is no duplication between them. In the second experiment, we used more complex datasets. The training set and the validation set are from the Collection dataset, and we use the Internet [17] as the test set.

Table 1 shows the results of this group of experiments. The results show that after using our proposed method, all indicators on the training set, validation set, and test set have been improved.

Table 1 Performance comparison with and without using our method for Dice loss.

| Method | Loss | Train&Val Dataset | Test Dataset | Val P(%) | Val J | Test P(%) | Test J |
|---|---|---|---|---|---|---|---|
| Gong *et al.*[34] | Dice loss | SBCoseg | SBCoseg | 99.0 | 0.981 | 99.1 | 0.949 |
| Modified framework | Dice loss + IS-Triplet loss | SBCoseg | SBCoseg | **99.4** | **0.985** | **99.2** | **0.951** |
| Gong *et al.*[34] | Dice loss | Collection | Internet | 98.7 | 0.978 | 94.5 | 0.760 |
| Modified framework | Dice loss + IS-Triplet loss | Collection | Internet | **99.3** | **0.987** | **95.0** | **0.801** |

Figure 5 shows the comparison charts of the change curve of loss, IOU, and precision on the training set and the validation set. It can be seen that after using our method, the Dice loss decreases faster, the validation IOU and Precision improve faster and better.

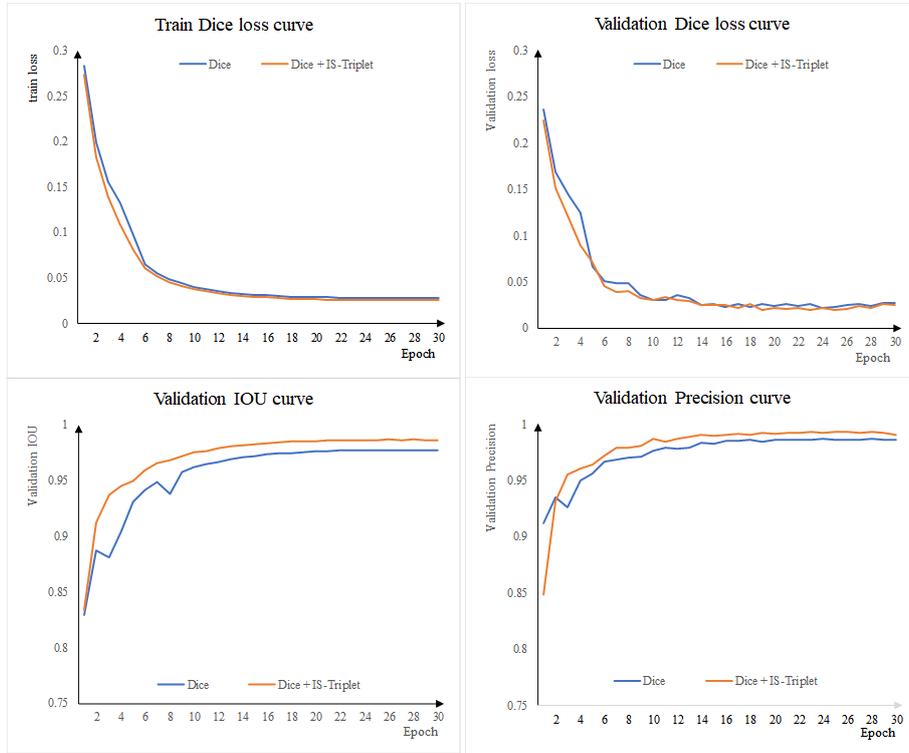

Fig. 4 The change curve of main indicators for using two methods on the Collection datasets.

## 4.3. The Second Group of Experiments

To further study the effects and applicability of our approach, we combine IS-Triplet loss with Dice loss, CE loss, and Focal loss. In this group of experiments, we use the Collection dataset as the training set and validation set, and Internet [17] as the test set.

Table 2 shows the results of this group of experiments. On the validation set, our metric learning method has an improved effect on all of the three loss functions. On the test set, our IS-Triplet loss improves the performance of Dice loss and CE loss but slightly decreases the performance of Focal loss.

Table2 The performance comparisons of combining our method with each of 3 losses respectively

| Serial | Method | Loss | Validation Precision | Validation Jaccard | Test mean Precision | Test mean Jaccard |
|---|---|---|---|---|---|---|
| 1 | Gong et al.[34] | Dice loss | 98.7 | 0.978 | 94.5 | 0.760 |
| 2 | Modified framework | Dice loss + IS-Triplet loss | **99.3** | **0.987** | **95.0** | **0.801** |
| 3 | Gong et al.[34] | CE loss | 99.2 | 0.985 | 94.5 | 0.789 |
| 4 | Modified framework | CE loss + IS-Triplet loss | **99.4** | **0.987** | **94.8** | **0.790** |
| 5 | Gong et al.[34] | Focal loss | 99.2 | 0.981 | **94.8** | **0.797** |
| 6 | Modified framework | Focal loss + IS-Triplet loss | **99.4** | **0.985** | 94.7 | 0.792 |

Figure 5 shows a comparison chart of the change curves of loss, IOU, and precision on the training set and the validation set as the training progresses. It can be seen from the figure that after using our method, the 3 loss functions decrease faster and better, and the IOU and Precision are also improved. Furthermore, although the original performance of these three loss functions is different, they can all be improved to almost the same level after using our metric learning method.

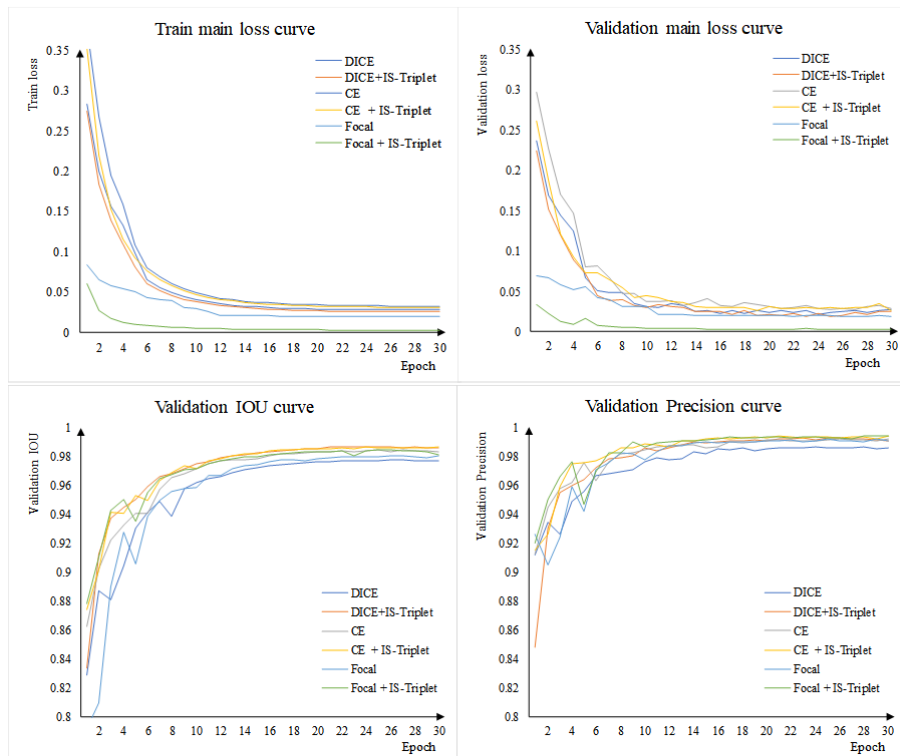

Fig.5 The change curve of main indicators when combining our method with each of 3 traditional loss functions on the Collection dataset.

# 5. Conclusions

In this paper, we propose a novel DML loss function for image co-segmentation, called IS-Triplet loss. By optimizing the multiple triples formed by the embedding features of pixels in the high-dimensional space, we can make the pixels of different categories have better distinguishability in high-dimensional space, thereby improving the distinguishability of pixels in the semantic space. To take into account the optimization effect and calculation speed, we also especially propose a targeted triple sampling strategy, which not only controls the upper limit of the sampling number of the triples but also eliminates the adverse effects caused by the imbalance of the foreground and the background points' numbers. We apply the proposed method to the image co-segmentation, combined with dice loss, CE loss, and Focal loss. The experiment results on the SBCoseg dataset and the Internet dataset show that after using our method, the traditional loss functions decrease faster and better, and the IOU and Precision are also improved. Furthermore, although the original performance of these 3 loss functions is different, they can all be improved to almost the same level after using our metric learning method.

In the future, we try to explore the application of the proposed approach to other image segmentation problems, such as semantic segmentation, video dynamic segmentation, etc.